\documentclass[10pt]{article}
\usepackage[a4paper,
            left=1in,
            right=1in,
            top=1in,
            bottom=1in,]{geometry}
\usepackage{amsmath,float}
\usepackage[authoryear,round]{natbib}
\usepackage{siunitx}
\usepackage{graphicx}  % For including graphics
\usepackage{caption}    % For customizing captions
\usepackage{subcaption} % For subfigures
\usepackage{amsmath}
\usepackage{tabularray}
\usepackage{bibentry}
\usepackage{tikz}
\usetikzlibrary{shapes.geometric, arrows, positioning}
\usepackage{graphicx}
\usepackage{color}
\graphicspath{ {images/} }
\title{\fontsize{16pt}{19.2pt}\selectfont \bf 
   Toward Transf\textit{OR}mers: Revolutionizing the Solution of Mixed Integer Programs with Transformers\thanks{A revised version of this paper has been accepted for publication in the 2024 IISE Conference Proceedings.}}
 \author{\fontsize{12pt}{14.4pt}
    Joshua F. Cooper,  Seung Jin Choi, and \.I. Esra B\"uy\"uktahtak{\i}n\thanks{email: esratoy@vt.edu, Grado Department of Industrial and Systems Engineering, Virginia Tech, Blacksburg, Virginia, USA}
 }%\date{\today}
\begin{document}
\maketitle
\begin{abstract}
In this study, we introduce an innovative deep learning framework that employs a transformer model to address the challenges of mixed-integer programs, specifically focusing on the Capacitated Lot Sizing Problem (CLSP). Our approach, to our knowledge, is the first to utilize transformers to predict the binary variables of a mixed-integer programming (MIP) problem. Specifically, our approach harnesses the encoder decoder transformer's ability to process sequential data, making it well-suited for predicting binary variables indicating production setup decisions in each period of the CLSP. This problem is inherently dynamic, and we need to handle sequential decision making under constraints. We present an efficient algorithm in which CLSP solutions are learned through a transformer neural network. The proposed post-processed transformer algorithm surpasses the state-of-the-art solver, CPLEX and Long Short-Term Memory (LSTM) in solution time, optimal gap, and percent infeasibility over 240K benchmark CLSP instances tested. After the ML model is trained, conducting inference on the model, reduces the MIP into a linear program (LP). This transforms the ML-based algorithm, combined with an LP solver, into a polynomial-time approximation algorithm to solve a well-known NP-Hard problem, with almost perfect solution quality.
\end{abstract}
\section{Introduction}
% \textcolor{red}{The title of the paper is very catchy. Shall we keep it for the journal article?}\textcolor{blue}{Thats probably best, perhaps somthing like: \\
% Towards Transf\textit{OR}mers: A first step in solving Mixed Integer Programs with Transformers}
% \textcolor{red}{I like the new title, which is more cautious about papers' contribution claims}
% \textcolor{red}{other contributors to the paper should be added as co-authors. Thanks!}
In the field of operations research, combinatorial optimization, particularly Mixed Integer Programming (MIP), represents a complex, yet crucial, area of study. The inherent NP-hard nature of MIP, especially in the context of capacitated lot size problems (CLSPs) that are NP-Complete, presents significant computational challenges \citep{nphard}. However, the emergence of deep learning (DL) algorithms offers a promising avenue to enhance solve times for such problems. By predicting binary variables, these algorithms can effectively reduce complex MIPs to linear programs that are more tractable \citep{lstm}.

%\subsection*{Gap Identification}
While the integration of machine learning (ML) into MIP has garnered considerable attention, and approaches that learn optimal solutions via integrated Long Short-Term Memory (LSTM)-optimization frameworks are gaining prominence, a consistent challenge persists: ensuring the feasibility of predicted solutions \citep{lstm}. The use of Sigmoid functions in output layers, akin to fixing in integer programming, often leads to suboptimal or infeasible solutions. This limitation is significant because it hinders the practical application of current ML techniques in scenarios where MIPs require rapid, repeated solving. Renowned solvers such as CPLEX and Gurobi, although efficient for many applications, fall short in speed for such demanding MIP scenarios, highlighting the need for ML methods that can bridge this gap without compromising the feasibility and optimality of the solution \citep{offtheshelf}. While others have proposed using a transformer architecture to solve MIPs,  these methods rely on reinforcement learning, graph representations, or both, making them less generalizable than a standard transformer \citep{TsTransformerSurvey}. 

%\subsection*{Research Focus}
This paper introduces the use of transformer models \citep{attentionisallyouneed}, a state-of-the-art deep learning approach, to address these challenges in dynamic mixed-integer programming. Transformers have already demonstrated superior performance in various machine learning applications, including natural language processing \citep{zeyer2019comparison, ezen-can2020comparison}, and computer vision \citep{LSTMvTransfomerQA}, often exceeding LSTM models in time series tasks \citep{TsTransformerSurvey}. This research aims to leverage their sequence-to-sequence (often called `Seq2Seq') capabilities to tackle sequential combinatorial optimization problems. 

%\subsection*{Objectives and Contributions}
The primary objective of this research is to advance the application of deep learning (DL) in mixed-integer programming by improving one or more of the following aspects over previous DL work: solve time, optimality gap, and infeasibility percentage \citep{lstm, yilmaz2023expandable}. By doing so, this study seeks to provide a more robust and efficient framework for solving complex MIPs, thereby enhancing the practical utility of machine learning in operations research.

\section{Methodology}
\subsection{CLSP: Mixed Integer Program Formulation}
The Capacitated Lot Size Problem (CLSP) is an optimization challenge that involves determining optimal production quantities for items over a series of time periods, subject to production capacity constraints, with the aim of minimizing total costs, including setup and inventory holding costs. Let $T$ represent the number of periods considered within the planning horizon, where $t \in \{1, 2, \ldots, T\}$ is the index that represents each period within the planning horizon. Let $d_t$, $p_t$, $f_t$, $h_t$, and $c_t$ represent the expected demand, the unit production cost, the setup cost, the unit inventory holding cost, and the production capacity in period $t$, respectively. Given that $x_t$, $s_t$, and $y_t$ are the variables that define the units produced, the ending inventory, and the binary production setup variable that is equal to 1 if there is production in period $t$, and 0 otherwise, in period $t$, the CLSP MIP formulation is presented as follows:
\begin{align}
    \min & \sum_{t=1}^{T} (p_t x_t + f_t y_t + h_t s_t) && \text{(Objective Function)} \label{eq1} \\
    \text{s.t.} \quad & s_{t-1} + x_t - d_t = s_t && \forall t = 1, 2, \ldots, T \label{eq2} \\
    & x_t \leq y_t c_t && \forall t = 1, 2, \ldots, T \label{eq3} \\
    & x_t, s_t \geq 0 && \forall t = 1, 2, \ldots, T \label{eq4} \\
    & y_t \in \{0, 1\} && \forall t = 1, 2, \ldots, T \label{eq5}
\end{align}
The model minimizes production, setup, and holding costs over periods \(t \in \{1,2,...,T\}\) via the objective function (1). Constraint (2) manages the inventory flow, ensuring that demand in period \(t\) is met by the prior inventory plus production. Constraint (3) caps production by capacity, incurring fixed costs for any production in \(t\). Constraint (4) mandates non-negative production and inventory levels. Finally, the constraint (5) sets \(y_t\) as binary, reflecting production decisions. The initial inventory is assumed to be zero.

\subsection{Design of the Transformer Model}
Our transformer model, inspired by \cite{attentionisallyouneed}, was experimented with various configurations to determine the optimal structure to solve CLSP. Although many model configurations were tested, the final configuration used in this paper is given in Table \ref{tab:modelConfig}.
\begin{table}[H]
\centering
\begin{tabular}{ll}
\hline
\textbf{Hyperparameters} &  \\ \hline
Parameter Count & 6.8 M\\
Source Vocabulary Size & 12000 \\
Target Vocabulary Size & 4 \\
Embedding Dimensions & 200 \\
Training GPU hours & $<1$ \\

\hline
\end{tabular}
\caption{Hyperparameter selection for the model.}
\label{tab:modelConfig}
\end{table}
While the architecture's overall structure adhered closely to the standard transformer model, extensive hyperparameter tuning was conducted to tailor the model for the specific nature of CLSPs.
Model training was predominantly successful with the teacher-forcing approach, in line with the findings of \cite{attentionisallyouneed}. This was attributed to the model's sensitivity to hyperparameters, where deviation from the specific set of parameters resulted in significantly prolonged convergence times. The optimality gap and infeasibility percentage, crucial metrics for evaluating the model's performance, were computed by running inference on a separate test set consisting of approximately 240,000 problem instances.

\subsection{Data and Preprocessing}
The model was trained with synthetically generated benchmark data following the CLSP instance generation schema of \cite{atamturk2004study}: capacity-to-demand ratios \( c \in \{3, 5, 8\} \), setup-to-holding cost ratios \( f \in \{1000, 10000\} \) and the number of periods \( T \in \{90\} \). The parameters regarding demand \( d_t \), unit production cost \( p_t \), production capacity \( c_t \), and setup cost \( f_t \) are generated from integer uniform distribution with the ranges, \( d_t \in [1,600] \), \( p_t \in [1,5] \), \( c_t \in [0.7c, 1.1c] \), \( f_t \in [0.9f, 1.1f] \), where \( c \) and \( f \) are defined in parameters. The inventory holding cost \( h_t \) is set to one.  CLSP instances were solved using CPLEX 22.1.0 and optimal solutions, objective function value, and solve times were stored in our dataset for validation.

The data set comprised 1,200,000 synthetic randomized instances of CLSP, generated and solved to create a complete training and testing dataset \citep{lstm}. These synthetic data provide a robust platform for the development and validation of models. 20\% of the data were used for validation and testing, the remaining data 80 \% making up the training data. Each six of the $c$ and $f$ combinations have 40,000 instances for validation and testing.
Pre-processing for the transformer involved batch normalization and log-scaling of the input data. Additionally, a vector embedding scheme was employed for numerical values similar to \cite{attentionisallyouneed}'s embedding scheme, but deviating from traditional word-focused embeddings, to better suit the numerical nature of CLSP data. The model presented was trained for 0.95 GPU hours.

\subsection{Post-Processing Transformer Predictions}
\label{Postprocessing Transformer Predictions}
\begin{figure}[h!]
\centering
\begin{tikzpicture}[node distance=1cm and 1cm]

% Define styles
\tikzstyle{process} = [rectangle, minimum width=2cm, minimum height=.75cm, text centered, draw=black, fill=orange!30]
\tikzstyle{decision} = [diamond, minimum width=2cm, minimum height=.75cm, text centered, draw=black, fill=green!30]
\tikzstyle{arrow} = [thick,->,>=stealth]

% Nodes
\node (input) [process] {CLSP Instance};
\node (transformer) [process, below of=input] {Transformer};
\node (clpex1) [process, below of=transformer, xshift=-2.5cm] {CPLEX: Flip Last Period};
\node (clpex2) [process, below of=transformer, xshift=2.5cm] {CPLEX: No Flip Last Period};
\node (feasibility) [process, below of=transformer, yshift=-1cm] {Check for Feasibility};
\node (solution) [process, below of=feasibility] {Solution};

% Lines
\draw [arrow] (input) -- (transformer);
\draw [arrow] (transformer) -| (clpex1);
\draw [arrow] (transformer) -| (clpex2);
\draw [arrow] (clpex1) |- (feasibility);
\draw [arrow] (clpex2) |- (feasibility);
\draw [arrow] (feasibility) -- (solution);

\end{tikzpicture}

\caption{Post-processing schema employed to remove infeasibility in predictions}
\label{fig:postprocessing}
\end{figure}
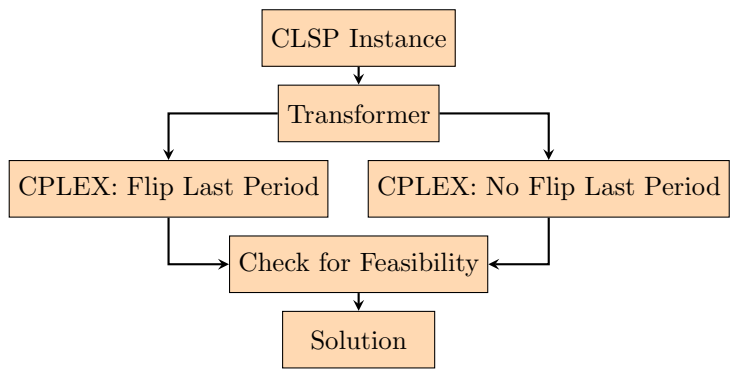

The predicted $y_t$-variables are fixed in the CLSP MIP model \eqref{eq1}--\eqref{eq5}, reducing the problem into an LP. Then the resulting LP is solved by CPLEX 22.1.0. Our initial model's prediction results were accurate, except for the prediction of the variable $y_t$ at time $T$ in 2\% of the test cases, as shown in Table \ref{tab:my_label}, where the column `Inf' gives the percentage of instances that are infeasibly predicted for each of the six combinations of $c$ and $f$.
\begin{table}[h!]
\centering
\begin{tabular}{lcc}
\hline
$c$ & $f$  & Inf (\%) \\ \hline
3 & 1000 & 4.444 \\
 & 10000 & 0.655 \\
5 & 1000 & 1.976 \\
 & 10000 & 0.101 \\
8 & 1000 & 1.191 \\
 & 10000 & 0.010 \\ \hline
\end{tabular}
\caption{Model infeasibility rate with no post-processing or $\langle EOS \rangle$ token.}
\label{tab:my_label}
\end{table} To address the imperfect predictions in the final period, we present a heuristic that runs two potential scenarios through CPLEX in parallel and selects only the feasible outcome, as demonstrated in Figure \ref{fig:postprocessing}. This approach empirically results in only optimal feasible solutions as output while significantly improving the solve time as discussed in detail in the next section. We also found that the use of an End of Sequence  $\langle EOS \rangle$ token eliminates the infeasiblity while incurring very marginally higher training time and lower inference time compared to our post-processing techniques. %\textcolor{blue}{Alternate approaches to rectifying this exist, but none solve the underlying failure to predict the last period perfectly.}

\section{Results}
\subsection{Model Performance}
The transformers model (TS) demonstrates exceptional performance in predicting the binary variables of the CLSP, as detailed in Table \ref{tab:avg_results}. The TS Solve Time' gives the combined CPU and GPU seconds to solve the CLSP instance with the integrated transformer framework (computational experiments were performed with an A100 GPU and 2 CPU cores while CLPEX solve times used 20 cores). The percentage of infeasibility (Inf \%) represents the proportion of infeasible solutions generated by the transformer model, while the optimality gap (Optgap \%) illustrates the percentage difference between the optimal solution obtained by CPLEX and the solution achieved by integrating the predictions of the transformer model with CPLEX. Our integrated transformer framework achieves an infeasibility percentage (Inf \%) and an optimality gap (Optgap \%) of 0\%, while significantly reducing CPLEX solve times by more than 99\%.
\begin{table}[h!]
\centering
\begin{tabular}{lccccc}
\hline
 C & $f$ & TS Solve Time & Optgap(\%) & Inf (\%) \\
\hline
3 & 1000 & 0.00294 & 0.000
 & 0.000 \\
 & 10000 & 0.00300 & 0.000 & 0.000 \\
5 & 1000 & 0.00318 & 0.000 & 0.000 \\
 & 10000 & 0.00312 & 0.000 & 0.000 \\
8 & 1000 & 0.00318 & 0.000 & 0.000 \\
 & 10000 & 0.00313 &0.000 & 0.000 \\
\hline
\end{tabular}
\caption{Summary of Average Computational Results with Post-Processing or $\langle EOS\rangle$ token.}
\label{tab:avg_results}
\end{table}

Noticeably, the initial model performs better with larger $f$ values (Figure \ref{tab:my_label}). This is probably a function of the frequency of production in the last period for each configuration in the data set.
    \label{fig:tsChart}
%\end{figure}

%Figure \ref{fig:tsChart} shows a number of the model's interesting behaviors. Notably, an initial stage when \( t < 15 \) where the model performs better before stabilizing at \( t = 15 \). After this point, the model has periodic fluctuations in accuracy with the linear regression having a  \( \hat{b}_1 \) of 0.0002, implying a capacity to generalize the predictions to longer sequences. 

\subsection{Comparative Analysis}
In this section, we compare our transformer-based prediction framework with the LSTM-based approach presented in \cite{lstm}. Our results reveal a significant improvement over the LSTM approach when the transformer model is trained and tested using CLSP instances with $T=90$ periods. In Table \ref{Table2}, TimeCPX represents the time that CPLEX solves the MIP model, TimeML gives the inference time for the ML model, Timegain (\%) presents the percent reduction in solve time compared to TimeCPX. Inf (\%) and Optgap (\%) represent the infeasibility rate and the optimality gap, respectively.

\begin{table}[h!]
\centering
\begin{tblr}{
  width = \linewidth,
  colspec = {Q[27]Q[50]Q[20]Q[30]Q[67]Q[67]Q[67]Q[69]Q[69]Q[67]Q[67]Q[67]Q[67]},
  cell{1}{3} = {c=1}{0.1\linewidth}, % Adjusted column span
  cell{1}{5} = {c=2}{0.1\linewidth},
  cell{1}{7} = {c=2}{0.1\linewidth},
  cell{1}{9} = {c=2}{0.1\linewidth},
  cell{1}{11} = {c=2}{0.1\linewidth},
  cell{1}{13} = {c=2}{0.1\linewidth},
  cell{3}{1} = {r},
  cell{3}{2} = {r},
  cell{4}{1} = {r},
  cell{4}{2} = {r},
  cell{5}{1} = {r},
  cell{5}{2} = {r},
  cell{6}{1} = {r},
  cell{6}{2} = {r},
  cell{7}{1} = {r},
  cell{7}{2} = {r},
  cell{8}{1} = {r},
  cell{8}{2} = {r},
  hline{1,9} = {-}{0.08em},
  hline{2} = {-}{},
  hline{10,9} = {-}{0.08em},
}
c & $f$     & TimeCPX &       & TimeML &        & Timegain(\%) &       & Inf(\%) &       & Optgap(\%) &       \\
  &       &          & & LSTM & TS               & LSTM & TS      & LSTM & TS      & LSTM & TS  \\
3 & 1000  &   0.107  &    & 0.02 & 0.003        & 81.31 & 97.20      & 0.00 & 0.00      & 0.59 & 0.00 \\
5 & 1000  &   0.07   &    & 0.02 & 0.003          & 71.43 & 95.71      & 1.17 & 0.00      & 0.16 & 0.00 \\
8 & 1000  &   0.06   &    & 0.02 & 0.003          & 66.67 & 95.00      & 0.00 & 0.00      & 0.05 & 0.00 \\
3 & 10000 &   0.79   &    & 0.02 & 0.003          & 97.46 & 99.62       & 67.70 & 0.00    & 0.80 & 0.00 \\
5 & 10000 &   0.65   &    & 0.02 & 0.003          & 96.92 & 99.54     & 42.00 & 0.00    & 1.30 & 0.00 \\
8 & 10000 &   0.33   &    & 0.02 & 0.003           & 93.93 & 99.09      & 22.60 & 0.00    & 3.18 & 0.00 \\
Average:&&  0.334    &      &0.02&0.003        & 94.02& 99.10        &22.245  &0.00   &1.013  &0.00

\end{tblr}
\caption{Performance comparison between LSTM and Transformer with $T=90$ period instances. LSTM results are taken from \cite{jinetal2024iise} using 10k instances per each $c$ and $f$ configuration.}
\label{Table2}

\label{tab: comparison}
\end{table}

%\begin{table}[h!]
%\centering
%\begin{tblr}{
  %width = \linewidth,
  %colspec = {Q[l]Q[c]Q[c]Q[c]Q[c]Q[c]Q[c]},
%}
%\hline
%c & $f$     & TimeGRB & Time\langle EOS\rangle & Timegain(\%) & Inf(\%) & Optgap(\%) \\
%\hline
%3 & 1000  & 0.06   & 0.003  & 95.00       & 0.00    & 0.00 \\
%5 & 1000  & 0.13    & 0.003  & 97.69       & 0.00    & 0.00 \\
%8 & 1000  & 0.10    & 0.003  & 97.00        & 0.00    & 0.00 \\
%3 & 10000 & 0.07    & 0.003  & 95.71        & 0.00    & 0.00 \\
%5 & 10000 & 0.12    & 0.003  & 97.50        & 0.00    & 0.00 \\
%8 & 10000 & 0.10    & 0.003  & 99.09        & 0.00    & 0.00 \\
%\hline
%Average & & 0.096  & 0.003  & 96.875        & 0.00    & 0.00 \\
%\hline
%\end{tblr}
%\caption{Performance comparison using Transformer (TS) with $T=90$ period instances using an \textlangle EOS\textrangle token. Results are using 10k instances per each $c$ and $f$ configuration. TimeGRB is solve time using Gurobi with the facet-defining $(\ell,S)$ inequalities proposed by \cite{barany1984strong}.}
%\label{Table2}
%\end{table}

\begin{table}[h!]
\centering
\begin{tabular}{
  >{\raggedright\arraybackslash}p{1cm} 
  >{\centering\arraybackslash}p{2cm} 
  >{\centering\arraybackslash}p{2cm} 
  >{\centering\arraybackslash}p{2cm} 
  >{\centering\arraybackslash}p{2cm} 
  >{\centering\arraybackslash}p{2cm} 
  >{\centering\arraybackslash}p{2cm}
}
\hline
$c$ & $f$ & TimeGRB & Time\textlangle EOS\textrangle & Timegain(\%) & Inf(\%) & Optgap(\%) \\
\hline
3 & 1000  & 0.06   & 0.003  & 95.00       & 0.00    & 0.00 \\
5 & 1000  & 0.13   & 0.003  & 97.69       & 0.00    & 0.00 \\
8 & 1000  & 0.10   & 0.003  & 97.00       & 0.00    & 0.00 \\
3 & 10000 & 0.07   & 0.003  & 95.71       & 0.00    & 0.00 \\
5 & 10000 & 0.12   & 0.003  & 97.50       & 0.00    & 0.00 \\
8 & 10000 & 0.10   & 0.003  & 99.09       & 0.00    & 0.00 \\
\hline
Average & & 0.096  & 0.003  & 96.875      & 0.00    & 0.00 \\
\hline
\end{tabular}
\caption{Performance comparison using Transformer (TS) with $T=90$ period instances using an \textlangle EOS\textrangle token. Results are using 10k instances per each $c$ and $f$ configuration. TimeGRB is solve time using Gurobi with the facet-defining $(\ell,S)$ inequalities proposed by \cite{barany1984strong}.}
\label{Table2}
\end{table}

The model outperforms CPLEX in terms of solve time with an improvement of more than 99\% on average and outperforms Gurobi with (l,S) inequalities by 96.9\%. The model is also superior to the LSTM in terms of solution quality, inference time, and training time.

As shown in Table \ref{Table2}, the post-processed transformer model achieved a higher time gain and a lower infeasibility percentage and optimality gap, finding the optimal solution each time, a notable achievement compared to the LSTM method's 22\% average infeasibility rate and 1\% optimality gap. While LSTM's inductive biases allow it to perform well on longer time horizons, the transformer model's performance makes it state of the art for in-sample problem lengths.

\subsection{Significance of Research Results}
As presented in the former section, the proposed post-processed transformer algorithm and transformer with $\langle EOS \rangle$ token surpasses the state-of-the-art solvers, CPLEX and Gurobi, and Long Short-Term Memory (LSTM) in solution time, optimal gap, and percent infeasibility over 240K benchmark CLSP instances tested. The results underscore the potential of transformer models to effectively solve mixed-integer programs, particularly in cases where rapid and repeated solving is crucial. Despite the fact that transformers have fewer inductive biases for sequential problems compared to LSTM architectures, they showed an exceptional ability to learn binary variables in MIPs. These findings position transformer models as a state-of-the-art approach to deep learning methods for MIPs, far surpassing traditional heuristics.
\section{Discussion and Future Work}
%\textcolor{red}{this is a big claim and we shall replace it with "efficient algortihm" as the training time for the transformer is not polynomial} \textcolor{blue}{JC: A. I was here referring to inference which is polynomial time as is shown in the appendix of \cite{attentionisallyouneed} B. all of the algoriths used to train the NN are also polynomial time. The only reason training is not strictly considered polynomial time is because of the stocastisity of SDG or ADAM and because these algo's don't guarantee optimally} %\textcolor{red}{YES, that is correct; ADAm is not polynomial time, its performance depends on the size and type of the instance}
We present an efficient algorithm in which CLSP solutions are learned through a transformer neural network. The proposed post-processed transformer algorithm achieves state-of-the-art performance in solution time, optimal gap, and percent infeasibility. After training the model, running inference with post-processing on in-distribution data relaxes the problem to a Linear Programming (LP) instance, solvable in polynomial time. This transformation effectively converts the learned algorithm into a polynomial-time solution for a well-known NP-Hard problem. The model finds the optimal solution in 100\% of the test cases when trained with an $\langle EOS \rangle$ token or simple post-processing is performed. 
%The performance variations across different transformer model configurations in solving the CLSP highlight an intriguing aspect of transformer scalability. Contrary to initial expectations, larger model sizes did not linearly correlate with better performance. Instead, the standout performance of the 17.3 M and 8.8 M parameter model suggests that extended training duration (figure \ref{fig:LossPlot}), rather than sheer model complexity, plays a critical role in achieving optimal results. This observation indicates that the problem complexity might not necessitate extremely large transformer models, and that sufficient training time is a key factor in model efficacy. 
\subsection{Implications for Operations Research}
Our findings have significant implications for the field of operations research, particularly in the realm of combinatorial optimization. The ability of transformer models to deliver fast, feasible, and near-optimal solutions in data-rich environments can revolutionize practical applications that demand efficient problem solving capabilities, such as online routing and logistics, energy management, and dynamic resource allocation. This study lays the groundwork for future research to focus on deep learning, especially transformers, as a viable approach to developing advanced heuristics that are based on deep learning to solve combinatorial optimization problems.

\subsection{Limitations}
Compared to traditional heuristics and other machine learning approaches, our transformer-based method demonstrates superior or equivalent performance, with the added benefit of faster training times. However, it is important to note that transformers are data intensive and require an in-depth understanding of deep learning for effective training and tuning. This limitation suggests that, while transformers offer considerable advantages, they are not a universal solution for all combinatorial optimization challenges. The finicky nature of these models could affect their broader applicability, as proper training and tuning are critical for optimal performance. This limitation underscores the need for expertise in deep learning to harness the full potential of transformer models in operations research applications. Additionally, the inductive biases of other architectures may be preferential for some MIPs, such as knapsack or traveling salesman problems. The base model fails to predict the last period of the CLSP in 2\% of instances without end of sequence token. Although the exact dynamics responsible for this behavior is unknown, positional bias is likely to be the culprit \citep{bias}.

\subsection{Future Work}
Future research will investigate testing the transformer model for longer-horizon CLSP problems and compare the results with LSTMs and expand the generalizability of the presented approach. The promising results obtained in this study advocate for the exploration of transformer applications in a wider range of combinatorial optimization problems. Future research should focus on testing the generalizability of our methods to other types of MIPs and developing more versatile models akin to advances in Natural Language Processing (NLP). Furthermore, solving the last-period prediction problem is a direction for further research.
%\nocite{*}

\bibliographystyle{plainnat}
\bibliography{refs}

%\appendix
%\section*{Appendix}
%\begin{figure}[H]
%\centering
%\includegraphics[scale=.4]{LossPlot.png}
%\caption{Training of the 2 Attention Head 2 layer transformer model showing loss vs. epochs}
%\label{fig:LossPlot}
%\centering
%\end{figure}

\end{document}